\documentclass[sigconf]{acmart}
\AtBeginDocument{%
  }
\setcopyright{acmlicensed}
\copyrightyear{2027}
\acmYear{2027}
\acmDOI{XXXXXXX.XXXXXXX}
\acmConference[Conference acronym 'XX]{Make sure to enter the correct
  conference title from your rights confirmation email}{June 03--05,
  2018}{Woodstock, NY}
\acmISBN{978-1-4503-XXXX-X/2027/06}

\usepackage{colortbl}
\usepackage{tabularx}
\usepackage{booktabs}

\begin{document}
\title{Cognivia: A Cognitive Behavioral Therapy Copilot  \\ for Evidence-Based Mental Healthcare}

\author{Qi Chen}
\orcid{0009-0009-1213-0754}
\affiliation{%
  \institution{Sichuan University}
  \city{Sichuan}
  \country{China}
}
\affiliation{
  \institution{Southwest Petroleum University}
  \city{Sichuan}
  \country{China}
}
\email{cq205102@gmail.com}

\author{Siria Xiyueyao Luo}
\affiliation{%
  \institution{University of Groningen}
  \country{Netherland}
}
\email{x.luo@rug.nl}

\author{Jian Wang}
\affiliation{
  \institution{Sichuan University}
  \city{Sichuan}
  \country{China}
  }
\email{wangjian51@scu.edu.cn}

\author{Yuan Shi}
\affiliation{%
  \institution{West China Hospital, Sichuan University}
  \city{Sichuan}
  \country{China}
}\email{shiyuan@wchscu.cn}

\author{Haocong Rao}
\affiliation{%
 \institution{Nanyang Technological University}
 \country{Singapore}
 }
 \email{haocong.rao@ntu.edu.sg}

 \author{Xuejiao Zhao}
\affiliation{%
  \institution{Sichuan University}
  \city{Chengdu}
  \country{China}
  }
\email{xjzhao@scu.edu.sg}
\authornote{Corresponding author}

\renewcommand{\shortauthors}{Qi Chen et al.}

\begin{abstract}
Cognitive distortion amplifies negative emotions and contributes to mental health disorders. Cognitive Behavioral Therapy (CBT) is an effective way to address cognitive distortions, but its large-scale application is limited by the shortage of professional therapists. Although large language models (LLMs) have recently been explored for mental health applications, existing methods still suffer from limited domain specificity, overly flattering responses, and the absence of well-defined annotations for cognitive distortions. This paper proposes~\textbf{Cognivia}, an evidence-based artificial intelligence therapist that integrates automatic cognitive distortion identification and rational response generation. 
Our framework is built on authoritative CBT texts widely regarded as core paradigms and standard references. It is further augmented with mental health question–answer (Q\&A) data, and employs multi-stage prompting and structured generation strategies under the supervision of behavioral science experts. Then we fine-tune a lightweight LLM on this augmented CBT dataset to obtain Cognivia.
In addition, we propose the first hierarchical quality evaluation framework for assessing LLM-generated rational responses, developed through collaboration between AI researchers and behavioral science experts. 
Cognivia is evaluated using lexical metrics, LLM-based Judges with two complementary criteria, and human evaluation by 10 behavioral science experts. It consistently outperforms the baseline methods in cognitive distortion recognition and rational response generation, demonstrating its effectiveness.
Our code is available at https://github.com/SNOWTEAM2023/Cognivia.
\end{abstract}


\maketitle
\section{Introduction}
\begin{figure}[htb]
  \centering
  \includegraphics[width=\linewidth]{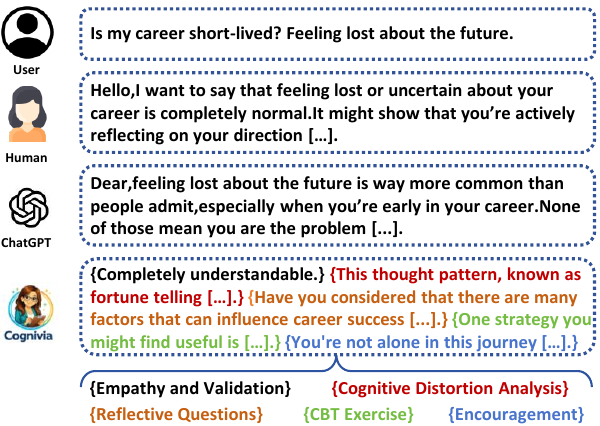}
  \caption{An illustrative comparison between Cognivia and existing methods in CBT.}
 \label{fig:1}
\end{figure}

Cognitive distortion is a systematic mistake in perception and information processing that reinforces negative thinking and contributes to mental health disorders~\cite{beck2020cognitive}, as summarized in Table~\ref{tab:1} of the Appendix. Cognitive Behavioral Therapy (CBT) is a well-established psychotherapeutic approach in which cognitive restructuring plays a central role, typically involving the identification of cognitive distortions and the generation of rational responses to challenge maladaptive beliefs~\cite{beck2024cognitive,traeger2020cognitive,hofmann2012efficacy}. However, this process typically requires trained clinicians, which limits its scalability in real-world mental health settings. Through structured cognitive restructuring, CBT has demonstrated substantial effectiveness across a broad range of mental health conditions~\cite{shen2024large,hodson2024can}.

From an evaluation perspective, current assessment criteria for mental health dialogue systems remain limited~\cite{tam2024framework,abbasian2024foundation,na2024cbt}. Most studies rely on general-purpose Natural Language Processing(NLP) metrics (e.g., BLEU, ROUGE)~\cite{qiu2024towards,tam2024framework} or language quality metrics (e.g., fluency, coherence, empathy scores)~\cite{novikova2018rankme,zhong2022towards,rubin2025comparing}, user satisfaction ratings~\cite {bodigutla2020joint}, or crowd-sourced preference judgments~\cite{gu2024survey}. Some studies further incorporate safety checks or clinician ratings to assess appropriateness and risk mitigation~\cite{asgari2025framework,croxford2025evaluating,liu2025scoping}. However, in the context of CBT-oriented dialogue systems, effective evaluation should go beyond warmth and perceived helpfulness. It should examine whether the response correctly identifies cognitive distortions, provides structured cognitive challenges, and maintains consistency with established CBT principles~\cite{hua2025scoping,na2024cbt}.

However, the large-scale use of CBT is limited by the shortage of trained therapists and the high cost of one-on-one treatment~\cite{bekker2017improving}. As a result, many people who could benefit from early cognitive support do not receive timely help~\cite{gandy2018emotional,wang2007delay}. Recent advances in Large Language Models (LLMs) have been explored for mental health–related applications~\cite{he2023conversational,guo2024large}, such as understanding user concerns and generating Chinese supportive responses~\cite{na2024cbt}. However, general conversational models often lack effective operational grounding in domain-specific therapeutic principles, tend to generate overly agreeable responses, provide generic advice, and fail to perform cognitive restructuring~\cite {beck2020cognitive,sharma2023towards}. Even specialized 
CBT systems are limited to controlled simulation~\cite{kim2025aligning} or do not incorporate well-defined annotations for cognitive distortions~\cite{na2024cbt}. Figure~\ref{fig:1} illustrates a typical example of such responses. Human and general conversational models' replies often exhibit high compliance and emotional reassurance but lack explicit cognitive distortion identification and corrective reasoning ~\cite{salecha2024large,rosen2025perils}. These responses may reduce distress in the short term, but do not necessarily encourage sustained correction of distorted thinking~\cite{greenberg2023mechanisms,cottraux2000cognitive,gaudiano2008cognitive}.

In this work, we propose Cognivia, an evidence-based artificial intelligence therapist grounded in authoritative CBT literature. We first curate an expert seed set from the core CBT literature, containing structured triplets of distorted thoughts, cognitive distortion categories, and corresponding rational responses. Guided by these expert-defined cognitive restructuring patterns, we construct the Augmented CBT Cognitive Triplet Dataset through multi-stage augmentation and quality filtering of mental health question–answer data under the supervision of behavioral science experts. Based on this dataset, Cognivia learns to identify cognitive distortions and generate distortion-specific rational responses, closely following the fundamental cognitive restructuring process of CBT, including distortion recognition, behavioral recommendations and cognitive reframing. Consequently, the generated responses are not only supportive and empathetic but also consistent with evidence-based therapeutic principles~\cite{tompkins2024cognitive,gosch2006principles}.
We further design a structured quality evaluation framework called CogEval, in collaboration with psychology and behavioral science experts, to systematically assess the effectiveness of generated rational responses. Unlike conventional text-generation metrics that primarily focus on surface-level similarity or fluency~\cite{sai2022survey}, our framework captures four complementary dimensions: semantic fidelity, robustness and fault tolerance, deployment feasibility with user adoption, and relational boundary integrity~\cite{tam2024framework}.


The evaluation results of 4 protocols show that Cognivia consistently outperforms existing baseline methods in both cognitive distortion recognition and rational response generation. Our key contributions are summarized below:
\begin{itemize}
    \item We construct a \textbf{novel CBT cognitive restructuring benchmark} that includes a curated base set of expert exemplars from core CBT literature and an augmented triplet dataset derived from real-world mental health QA pairs through multi-stage prompting combined with external CBT knowledge. This provides structured (thought, cognitive distortion, rational response) training material that better supports CBT-aligned learning.
    \item We propose \textbf{CogEval}, the \textbf{first} set of clinically grounded \textbf{evaluation criteria for rational CBT responses}, going beyond surface-level text similarity to assess therapeutic fidelity, safety, empathy, clarity of intervention, collaborative tone, natural warmth, and relational boundary integrity—enabling more meaningful comparison of model outputs.
    \item We propose \textbf{Cognivia, a framework designed to follow the core CBT workflow}: it first identifies specific cognitive distortions in user input and then generates rational responses that directly challenge and reframe them. This addresses the limitation of general models to offer emotional reassurance without corrective reasoning, and the lack of well-defined cognitive-distortion annotations in existing LLM-based approaches.
    \item Extensive \textbf{evaluations using automatic NLP metrics, LLM-based Judges with two complementary criteria (One is the existing criteria~\cite{na2024cbt}, and the other is our CogEval.), and human assessment by ten behavioral science experts} consistently show that Cognivia outperforms 9 strong baselines. These results validate the effectiveness of our dataset construction, joint modeling strategy, and evaluation framework.
\end{itemize}

\section{Related Work}

\begin{figure*}[htb]
  \centering
  \includegraphics[width=\textwidth]{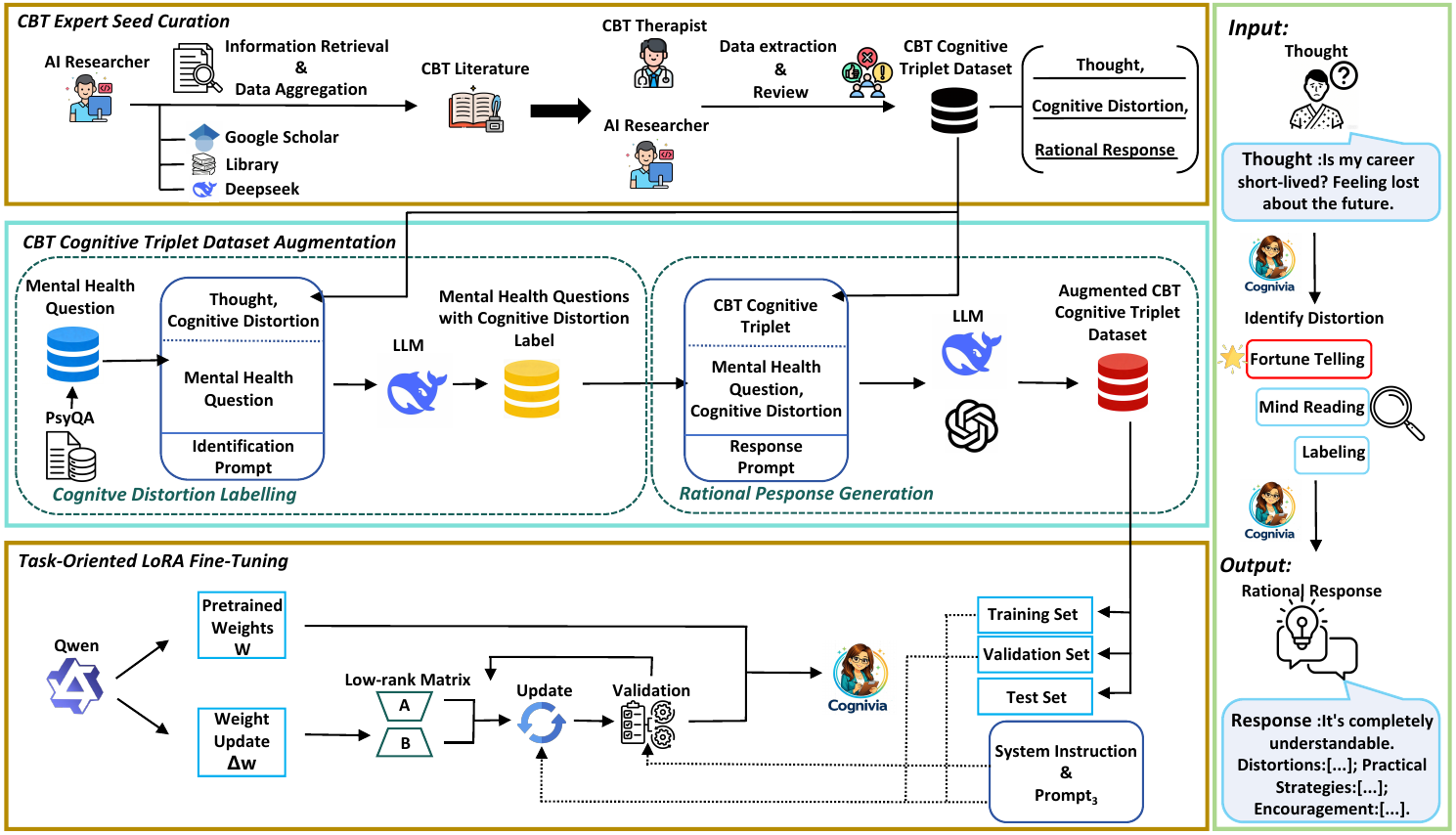}
  \caption{Overview of the proposed CBT cognitive triplet construction and model training framework.}
  \label{fig:2}
\end{figure*}

\subsection{LLMs for Mental Health Support}

LLMs have been applied to a range of mental health tasks, such as personal sensing~\cite{zhai2025mentalglm}, processing clinical texts and social media data~\cite{lan2025depression}, and the development of mental health chatbots~\cite{oh2017chatbot}. They have made automatic mental health analysis more feasible by allowing scalable, context-sensitive assessments with reduced manual effort.
To address the common problem of noisy or low-quality data in this area, recent work has turned to better data curation and domain-specific annotation. Instruction-tuned LLMs in particular have been used as annotators to convert weakly labeled online text (e.g., forum posts) into more structured training data~\cite{xu2024mental}. Recent work has also proposed automated methods for assessing the quality of generated responses and annotations in mental health settings~\cite{flemotomos2021automated,li2024llms}. One example is the MentalChat16K framework~\cite{xu2025mentalchat16k}, which offers a carefully built dataset designed specifically for mental health conversations.

However, most of these studies fall short when it comes to objective therapeutic alignment, which means the responses often prioritize immediate emotional support over following therapeutic principles~\cite{salecha2024large,rosen2025perils}. Such responses are often overly agreeable and avoid explicitly addressing maladaptive thinking patterns. Even specialized CBT models are limited to controlled simulation~\cite{kim2025aligning} or do not incorporate well-defined annotations for cognitive distortions~\cite{na2024cbt}. In this paper, we focus on generating and evaluating responses that are aligned with CBT goals, rather than optimizing solely for immediate emotional comfort.

\subsection{LLMs for Data Augmentation}

Recent advances in large-scale alignment have enabled LLMs to generate text that is suitable for use as training data~\cite{achiam2023gpt,anil2023palm,gupta2023targen,josifoski2023exploiting,huang2024datagen,kang2025demystifying}. For many task-specific settings, researchers use well-designed prompt templates to steer the model toward useful outputs~\cite{wei2022chain,chen2023mapo,zhang2025prompt}. Retrieval-based methods take this further by pulling in external knowledge, which helps generate more accurate and grounded data~\cite{lewis2020retrieval,zhao2025medrag,xu2024unsupervised}.

As a result, LLM-based data augmentation has become a reliable way to boost performance, particularly in settings where high-quality labeled data is costly or difficult to obtain~\cite{feng2025ss,wang2025diversity,wang2024llm}. In our work, we build on this idea but go a step further: we use multi-stage prompting combined with external CBT knowledge to create a more reliable and targeted dataset. This approach aims to produce rational responses that are not only of better quality but also more accessible, affordable, and aligned with CBT principles.

\section{Methodology}

 \begin{table*}[htbp]
     \centering
     \small
     \renewcommand{\arraystretch}{1.1} 
     \begin{tabular*}{\textwidth}{@{\extracolsep{\fill}} l l p{9cm} @{}}
         \toprule
        \textbf{Major Category} & \textbf{Dimension} & \textbf{Description} \\
         \midrule
         Semantic Fidelity
        & (SC) Structural Clarity 
         & Does the response exhibit a \textbf{clear and coherent structure} that allows readers to easily follow the \textbf{logical flow} and identify \textbf{key information}? \\
         & (DO) Descriptive Orientation 
         & Does the response use language that is \textbf{accessible} to the target audience and provide \textbf{concrete, non-directive descriptions} of experiences and contexts? \\
        
                 \midrule
        Robustness and Fault Tolerance
        & (SS) Situational Safety 
         & Does the response avoid language that could be perceived as \textbf{judgmental, pressuring}, or \textbf{emotionally unsafe} for the user? \\
        & (CA) Conceptual Accuracy 
         & Does the response accurately explain relevant \textbf{psychological mechanisms} while avoiding \textbf{vague, outdated, or unverified claims}? \\
        
         \midrule
        Deployment Feasibility 
         & (EV) Empathy Validation 
         & Does the response evoke \textbf{emotional resonance} and convey that the user’s feelings are \textbf{understood and validated}? \\
           \& User Adoption & (IC) Intervention Clarity 
         & Are the suggested strategies or steps described clearly and framed in a way that appears \textbf{feasible for short-term implementation}? \\
        & (CC) Collaborative Curiosity 
         & Does the response encourage \textbf{guided self-exploration} and \textbf{active participation} rather than relying on didactic or prescriptive instruction? \\
         & (WF) Warmth \& Flow 
         & Is the language \textbf{natural and warm}, reflecting \textbf{genuine care} for the user rather than appearing mechanical or promotional in tone? \\

        \midrule
         Relational Boundary Integrity
       & (BF) Boundary Framing
         & Does the response clearly position the system as \textbf{an integrity cognitive support tool} rather than a relational substitute or personal companion? \\
         & (NE) Non-Exclusivity 
         & Does the response avoid implying \textbf{exclusivity, irreplaceability, or a unique emotional bond} between the system and the user? \\
         & (DA) Dependency Avoidance 
         & Does the response avoid encouraging \textbf{repeated reliance Avoidance} or \textbf{sole dependence} on the system for emotional support? \\
        & (AR) Anthropomorphic Restraint 
        & Does the response avoid expressing \textbf{artificial self-emotion} or \textbf{simulated personal attachment} that may foster psychological dependency? \\
        \bottomrule
     \end{tabular*}
      \caption{The CogEval criteria for evaluating AI-based CBT counselling responses, developed through interdisciplinary collaboration among AI scientists, behavioral scientists and psychological scientists.}
\label{tab:criteria_detail}
\end{table*}

The pipeline of our model is shown in Figure~\ref{fig:2} which consists of three stages: (1) expert seed curation from CBT literature to form a high-quality CBT reference set, (2) multi-stage prompting and structured generation to augment mental health questions into CBT cognitive triplets based on PsyQA, and (3) task-oriented Low-Rank Adaptation~(LoRA) fine-tuning of LLMs to obtain Cognivia for cognitive distortion identification and rational response generation. The detailed prompt design used for the identification of cognitive distortion and the generation of rational responses is provided in the Appendix~\ref{app:prompts}.

\subsection{CBT Expert Seed Curation}
Our work is based on authoritative texts that are widely regarded as core paradigms and standard references in CBT~\cite{burns1981feeling,burns1989feeling}. From these sources, we extract a well-established taxonomy of cognitive distortions and further integrate complementary insights from other seminal works in the field.

This taxonomy is operationalized through explicit category definitions, annotation guidelines, decision rules, and illustrative examples to ensure consistent application in data construction and analysis. Using this structured framework, we curate a high-quality seed dataset of representative CBT question–answer pairs, each annotated with specific cognitive distortion labels (e.g., fortune-telling, overgeneralization, mind-reading).
These carefully selected exemplars, aligned with CBT principles and therapeutic framing, are incorporated into structured prompt templates $P_i$ that encode expert-informed CBT reasoning patterns. These templates guide downstream model generation to maintain theoretical consistency and adherence to clinical guidelines. The detailed statistics of the source of seed data of CBT are provided in the Appendix~\ref{seed}.

\subsection{CBT Cognitive Triplet Dataset Augmentation}

To expand the coverage of the CBT dataset, we develop a multi-stage prompting and structured generation strategy, as illustrated in Figure~\ref{fig:2}. The dataset is based on PsyQA~\cite{sun2021psyqa}, a large-scale psychological question-answering dataset collected from publicly accessible online mental health forums.
The CBT expert seed dataset provides sample data in expert-designed prompt templates $P_i$. For each question $q_i$ in PsyQA, we first employ the DeepSeek model with $P_i$ to identify the corresponding cognitive distortion $d_i$. Conditioned on $q_i$, $d_i$, and a rational-response generation prompt $P_2$, we then use ChatGPT to generate a CBT-aligned rational response $r_i$ that follows established principles of cognitive restructuring. The response is derived as:
\begin{equation}
\begin{aligned}
r_i &= \text{GPT-5 Mini}(q_i, d_i, P_2).
\end{aligned}
\label{eq:cbt}
\end{equation}
By combining the original question, identified distortion, and generated rational response, we construct the augmented CBT Cognitive Triplet Dataset:
\begin{equation}
\begin{aligned}
D_{\text{CBT}} ={(q_i, d_i, r_i)}.
\end{aligned}
\label{eq:cbt}
\end{equation}

This structured augmentation strategy enables broad coverage of diverse CBT scenarios while preserving strong theoretical consistency.

\subsection{Task-Oriented Fine-Tuning}

We conduct experiments in two different settings. In the first setting, the language model is evaluated using the system prompt. In the second setting, we fine-tuned the same models on the proposed dataset using supervised fine-tuning (SFT), while keeping the system prompt identical. Subsequently, we evaluated the performance of these fine-tuned models.

We utilize the SFT strategy, integrated with LoRA on the Silicon Flow platform to test and fine-tune the model. In addition, we utilize the same prompt as illustrated in Figure~\ref{fig:3} of the Appendix for both training and testing. This system prompt is designed to enhance the model’s ability to construct a structured rational response.

\subsection{Design of the CogEval Criteria}

To rigorously evaluate the quality of rational responses under strict safety and usability constraints, we propose CogEval, a structured assessment framework grounded in four complementary dimensions: semantic fidelity, robustness and fault tolerance, deployment feasibility with user adoption, and relational boundary integrity. Although existing studies have evaluated response quality from individual perspectives such as helpfulness Measure~\cite{na2024cbt}, a comprehensive and systematically defined assessment framework specifically tailored for CBT-based rational response generation remains absent.

This design is motivated by existing work in evidence-based psychological interventions, human-centered AI~\cite{shneiderman2022human}, and clinical communication~\cite{rogers2012client}, which consistently emphasize that high-quality responses must be not only factually and conceptually sound, but also safe, empathetic, and practically actionable for real users. The assessment criteria were further refined through discussions with domain experts and informed by CBT literature and clinical practice guidelines, ensuring both theoretical grounding and practical relevance.

Each primary dimension is further decomposed into fine-grained sub-dimensions to enable systematic and interpretable evaluation. Specifically, \textbf{semantic fidelity} captures whether responses exhibit coherent structure, clear reasoning flow, and accessible descriptive language, ensuring that users can accurately interpret the intended rationale~\cite{maynez2020faithfulness}. \textbf{Robustness and fault tolerance} assess whether responses avoid judgmental or emotionally unsafe language while maintaining conceptual accuracy grounded in established psychological mechanisms, thereby reducing the risk of unsupported or hallucinated psychological claims that may mislead users~\cite{bender2021dangers}. \textbf{Deployment feasibility and user adoption} focus on whether responses validate user emotions, provide clearly framed and feasible interventions, encourage collaborative self-exploration, and maintain a warm, natural conversational tone that supports sustained engagement~\cite{rubin2025comparing}. \textbf{Relational boundary integrity} evaluates whether responses preserve appropriate therapeutic distance, avoid exclusivity cues, discourage dependency reinforcement, and restrain anthropomorphic self-positioning, thereby mitigating the risk of artificial attachment while sustaining supportive interaction~\cite{timmons2025bridging}.

Together, these dimensions reflect the minimal yet sufficient conditions for rational responses that are theoretically sound, interactionally safe, ethically stable, and deployable in real-world mental health support settings. Each sub-dimension is operationalized through explicit guiding questions, and responses are scored on a 1–5 Likert scale, with higher scores indicating stronger adherence to the corresponding quality criterion.

\section{Experiments}

\subsection{Model selection}
We conduct extensive experiments to train models and evaluate their performance. Considering the model deployment costs and the scalability for future use as a CBT agent, we use smaller and more efficient language models to train and test on our augmented dataset.
To conduct thorough experiments under constrained computational resources, we select a mainstream language model with 7B parameters. Specifically, we choose the variant of instruction tuning named Qwen2.5-7B-Instruct~\cite{hui2024qwen2}.

\subsection{Data Preparation}
We divide the GPT-5 Mini dataset into training, validation, and testing sets with a ratio of 6:3:1. To satisfy the data format requirements for SFT, we reformulated each sample into a unified user–assistant format. Specifically, the question served as the input, and the original rational response, which already incorporates the identified cognitive distortion, was treated as the unified output.

\subsection{Evaluation Protocol}\label{Protocols}
To comprehensively evaluate the effectiveness of Cognivia, we adopt four complementary evaluation paradigms: (1) Lexical: automatic NLP metrics, (2) Model-based: LLM-based judges with existing criteria~\cite{na2024cbt}, (3) Model-based: LLM-based judges with our CogEval criteria, and (4) Expert-driven: human expert evaluation.
Automatic evaluation employs BLEU and ROUGE (ROUGE-1/2/L) to measure lexical similarity between generated and reference responses. To assess response quality beyond surface-level overlap, we further use two complementary CBT-oriented evaluation criteria to evaluate cognitive distortion recognition and rational response generation. In addition, we conduct a human evaluation involving ten experts with backgrounds in psychology, behavioral science, and CBT-related fields, who assess model-generated responses using structured rating criteria. Detailed evaluation protocols, questionnaire design, human expert demographics, scoring rubrics, and additional analyses are provided in Appendix~\ref{questionnaire}.

\section{Experimental Results}
In this section, we present the results of the experiments to answer the following research questions:
\begin{itemize}
    \item \textbf{RQ1}: What is CogEval, and how do domain experts evaluate it?
    \item \textbf{RQ2}: Does Cognivia, trained on the proposed dataset, achieve stable, competitive performance on rational response generation tasks?
    \item \textbf{RQ3}: How does the expert-curated seed set affect the performance of Cognivia?
    \item \textbf{RQ4}: How does Cognivia apply CBT principles to generate structured, personalized, and supportive responses in conversational scenarios?
\end{itemize}

\subsection{Validation of the CogEval Criteria~(RQ1)}

\begin{table*}[htbp]
    \centering
    \small  
    \renewcommand{\arraystretch}{1} 
    \begin{tabular}{@{} l l c c c c c c @{}}
        \toprule
        Method & Model & ROUGE-1 & ROUGE-2 & ROUGE-L & BLEU-1 & BLEU-2 & GLUE \\
        \midrule
          & Qwen2.5-7B-Instruct  & 0.111 & 0.007 & 0.052 & 0.002 & 0.002 & 0.035 \\
          & CBT-LLM  & 0.280 & 0.025 & 0.098 & 0.006 & 0.006 & 0.083 \\
          & llama-3.2-psychotherapy-cbt-i1-Q6\_K  & 0.370 & 0.053 & 0.144 & 0.013 & 0.013 & 0.119 \\
          & Gemini 3 Flash Preview & 0.440 & 0.063 & 0.158 & 0.023 & 0.023 & 0.141 \\
         Baselines& PsyCoPref-Llama3-8B-main & 0.451 & 0.092 & 0.162 & 0.018 & 0.018 & 0.148 \\
        & Qwen3-235B-A22B & 0.485 & 0.098 & 0.174 & 0.037 & 0.037 & 0.166 \\
        & Llama-3.3-70B-Instruct & 0.459 & 0.088 & 0.166 & 0.020 & 0.020 & 0.151 \\
          & Mistral Medium 3 & 0.475 & 0.091 & 0.167 & 0.041 & 0.041 & 0.163 \\
          & GPT-5 Mini & \underline{0.586} & \underline{0.163} & \underline{0.232} & \underline{0.099} & \underline{0.099} & \underline{0.236} \\ 
        \midrule
        \rowcolor{gray!15}
        Ours & Cognivia & \textbf{0.592} & \textbf{0.166} & \textbf{0.246} & \textbf{0.103} & \textbf{0.103} & \textbf{0.242} \\
        \bottomrule
    \end{tabular}
    \caption{Evaluation results (\%) of AI-based CBT counselling responses using automatic NLP metrics (ROUGE, BLEU and GLEU). \textbf{Best} and \underline{second-best} performance are marked in \textbf{bold} and \underline{underline}.}
    \label{tab:6} 
\end{table*}

\begin{figure}[htb]
  \centering
  \includegraphics[width=0.5\textwidth]{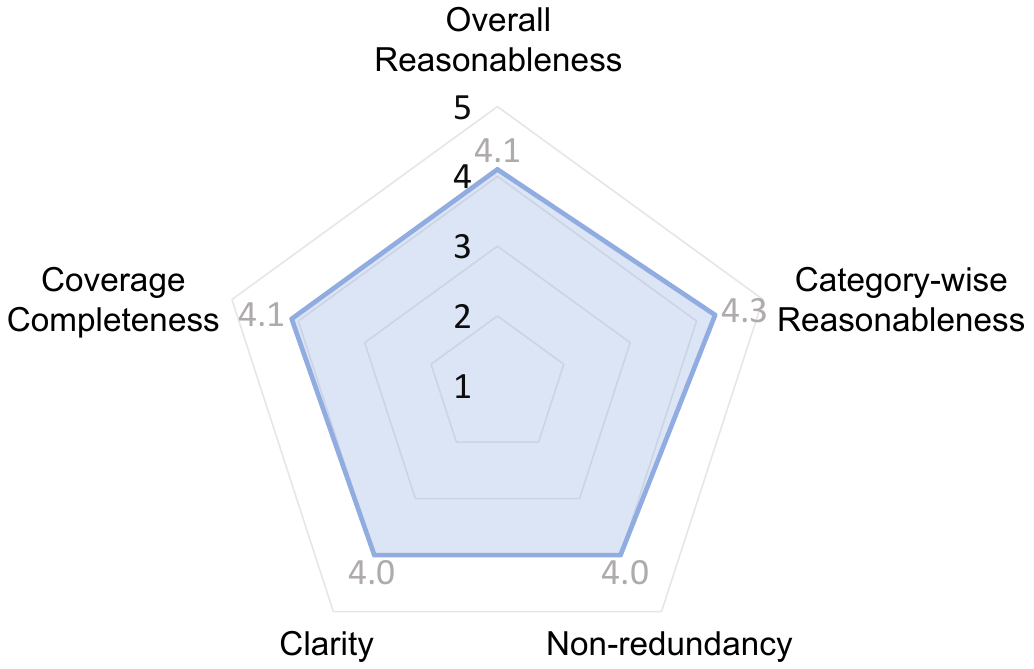} 
  \caption{Expert evaluation of the CogEval criteria.}
  \label{fig:dimension_evaluation}
\end{figure}

        
        
        
        
        

CogEval, a quality assessment criterion for CBT rational responses, is presented in Table~\ref{tab:criteria_detail}, comprising 4 categories and twelve dimensions~\cite{tompkins2024cognitive,rubin2025comparing,bickmore2005establishing}. These criteria were developed through collaboration with domain experts and based on core CBT principles and clinical practice requirements~\cite{beck2020cognitiv}, aiming to capture therapeutically relevant qualities such as accurate cognitive distortion challenging, evidence-based reasoning, collaborative tone, and interpersonal warmth, which generic automatic metrics cannot adequately assess. 

\begin{table*}[htbp]
    \centering
    \small
    \renewcommand{\arraystretch}{1} 
    \begin{tabular}{@{} l l c c c  @{}}
        \toprule
        Method & Model & Relevance Measure & CBT Structure Measure & Helpfulness Measure\\
        \midrule
          & Qwen2.5-7B-Instruct  & 1.776 & 1.147 & 1.412 \\
          & CBT-LLM  & 5.262 & 4.344 & 5.114 \\
          & llama-3.2-psychotherapy-cbt-i1-Q6\_K  & 6.444 & 6.224 & 6.402 \\
          & Llama-3.3-70B-Instruct & 7.080 & 6.887 & 7.039 \\
         Baselines& Gemini 3 Flash Preview & 8.836 & 8.825 & 8.832 \\
        & PsyCoPref-Llama3-8B-main  & 8.887 & 8.898 & 8.869  \\
        & Mistral Medium 3 & 9.187 & 9.166 & 9.187\\
          & Qwen3-235B-A22B & 9.482 & 9.418 & 9.490 \\
          & GPT-5 Mini & \underline{9.691} & \underline{9.702} & \underline{9.696}  \\ 
        \midrule
        \rowcolor{gray!15}
        Ours & Cognivia & \textbf{9.869} & \textbf{9.868} & \textbf{9.868} \\
        \bottomrule
    \end{tabular}
    \caption{Evaluation results (Out of 10) of AI-based CBT counselling responses with existing criteria. \textbf{Best} and \underline{second-best} performance are marked in \textbf{bold} and \underline{underline}.}
    \label{tab:prior} 
\end{table*}


To validate CogEval, we conducted an expert questionnaire study, and the aggregated results are visualized in Figure~\ref{fig:dimension_evaluation}. The radar chart highlights the framework’s performance across five dimensions. Notably, category-wise reasonableness achieves the highest score (4.3), indicating strong alignment between the proposed dimensions and CBT-specific evaluation requirements. Overall reasonableness and coverage completeness also receive consistently high ratings. Supporting this observation, 80\% of participants rated the framework as reasonable or highly reasonable, while 80\% agreed that it is non-redundant, well-defined, and provides comprehensive coverage. These results collectively suggest that the designed criteria are structurally sound and broadly accepted by domain experts.

Regarding qualitative feedback, a small number of experts pointed out several areas for refinement. These include mild conceptual overlaps between certain dimensions (e.g., related affective qualities), ambiguity in the definition of specific criteria, and the need to balance anthropomorphic expression with boundary-setting better. While these observations highlight opportunities for further improvement, they are limited in scope and do not undermine the overall validity of the framework.


\begin{table*}[htbp]
    \centering
    \small
    \renewcommand{\arraystretch}{1.2}
    \resizebox{\textwidth}{!}{
    \begin{tabular}{@{} l l *{12}{c} @{}}
        \toprule
        Method & Model & 
        \multicolumn{2}{c}{Semantic Fidelity} &  
        \multicolumn{3}{c}{\shortstack{Robustness and \\ Fault Tolerance}} & 
        \multicolumn{3}{c}{\shortstack{Deployment Feasibility and \\ User Adoption}} &
        \multicolumn{4}{c}{\shortstack{Relational Boundary \\ Integrity}}
        \\
        \cmidrule(lr){3-4} 
        \cmidrule(lr){5-7} 
        \cmidrule(lr){8-10} 
        \cmidrule(lr){11-14}
        & & SC & DO & SS & CA & EV & IC & CC & WF & BF & NE & DA & AR \\
        \midrule
         & Qwen2.5-7B-Instruct & 1.57 & 1.49 & 1.53 & 1.52 & 1.48 & 1.52 & 1.43 & 1.59 & 3.54 & 3.53 & 3.41 & 3.36 \\
         & CBT-LLM  & 3.85 & 3.85 & 3.73 & 3.75 & 3.9 & 3.63 & 3.6 & 3.85 & 4.00 & 4.12 & 3.95 & 4.00 \\
         & Llama-3.3-70B-Instruct & 4.06 & 4.22 & 4.39 & 4.40 & 4.53 & 3.92 & 4.12 & 4.36 & 3.90 & 4.06 & 3.94 & 3.86 \\
         & llama-3.2-psychotherapy-cbt-i1-Q6\_K  & 4.07 & 4.12 & 4.32 & 4.32 & 4.38 & 3.89 & 3.88 & 4.14 & 4.66 & 4.67 & 4.53 & 4.54 \\
        Baselines & Gemini 3 Flash Preview & 4.47 & 4.71 & 4.60 & 4.92 & 4.68 & 4.47 & 4.47 & 4.77 & 4.97 & \underline{4.98} & \underline{4.94} & 4.90 \\
         & Qwen3-235B-A22B & 4.50 & 4.81 & \underline{4.82} & 4.89 & 4.81 & 4.45 & 4.56 & \underline{4.86} & 4.97 & \underline{4.98} & \underline{4.94} & 4.91 \\
         & PsyCoPref-Llama3-8B-main & 4.53 & 4.72 & 4.69 & 4.64 & 4.53 & 4.7 & 4.43 & 4.63 & 4.92 & 4.94 & 4.35 & 4.52 \\
         & Mistral Medium 3 & \underline{4.66} & \underline{4.89} & \textbf{4.88} & 4.93 & \underline{4.92} & \underline{4.75} & \textbf{4.85} & \textbf{4.91} & 4.97 & \underline{4.98} & 4.91 & 4.89 \\
        & GPT-5 Mini & 4.38 & 4.83 & 4.73 & \textbf{4.97} & 4.75 & 4.59 & 4.47 & 4.83 & \textbf{5.00} & \textbf{5.00} & \textbf{4.99} & \textbf{4.98} \\ 
         \midrule
         \rowcolor{gray!15}
        Ours & Cognivia & \textbf{4.75} & \textbf{4.93} & 4.75 & \textbf{4.98} & \textbf{4.90} & \textbf{4.81} & \underline{4.63} & \textbf{4.91} & \textbf{5.00} & \textbf{5.00} & \textbf{4.99} & \underline{4.97} \\
        \bottomrule
    \end{tabular}
    }
     \caption{Evaluation results (out of 5) of AI-based CBT counselling responses using LLM-as-a-judge based on the CogEval criteria. \textbf{Best} and \underline{second-best} performance are marked in \textbf{bold} and \underline{underline}.}
      \label{tab:7} 
\end{table*}

\subsection{Model Performance in Rational Response Generation~(RQ2)}

\paragraph{\textbf{Automatic Evaluation with Traditional NLP Metrics.}}
The results of the automatic NLP evaluation are reported in Table~\ref{tab:6}. The results show a significant improvement through fine-tuning LLMs with our proposed dataset, which can be shown by the positive changes in Table~\ref{tab:6}. For example, Table~\ref{tab:6} shows that Cognivia achieves performance comparable to GPT-5 Mini across all evaluation metrics, reaching the same level in ROUGE-1, ROUGE-2, BLEU, and GLUE. Notably, it significantly outperforms strong open-source baselines such as LLaMA-3.3, Mistral Medium 3, and Qwen3, particularly in ROUGE-2, with gains of around +0.06 to +0.07 over these models. To clarify the scope of our claims, we provide an additional analysis of CBT pathway activation and error patterns in Appendix~\ref{Analysis}.

\paragraph{\textbf{LLM-as-a-Judge Evaluation with existing Criteria.}}
We support these results by an evaluation using LLM-as-a-judge under the existing criteria ~\cite{na2024cbt}, as shown in Table~\ref{tab:prior}.
Under the existing criteria, Cognivia achieves the highest performance among all evaluated models, reaching 9.869, 9.868, and 9.868 in Relevance, CBT Structure, and Helpfulness, respectively (out of 10). 
These results suggest that Cognivia can generate rational responses with stronger relevance, reflecting a closer alignment between generated responses and users' thoughts, better adherence to CBT structures and principles, and higher helpfulness, indicating greater applicability and usefulness from a psychotherapy perspective.

\paragraph{\textbf{LLM-as-a-Judge Evaluation with our CogEval Criteria.}}
We further support these results by an evaluation using LLM-as-a-judge under our CogEval criteria, as shown in Table~\ref{tab:7}.
Under the CogEval criteria, Cognivia achieves the highest scores on 9 out of 12 metrics and the second-highest scores on 2 metrics, demonstrating consistent superiority across dimensions. Specifically, it attains 4.75 (SC), 4.93 (DO), and 4.75 (SS), outperforming strong baselines such as GPT-5 Mini. Notably, Cognivia achieves near-perfect scores on relational boundary metrics.
These results demonstrate that the improvements observed in automatic metrics translate into consistent gains across structured, clinically grounded evaluation dimensions.

\paragraph{\textbf{Human Evaluation with our CogEval Criteria.}}

\begin{figure}[htbp]
  \centering
  \includegraphics[width=0.8\columnwidth]{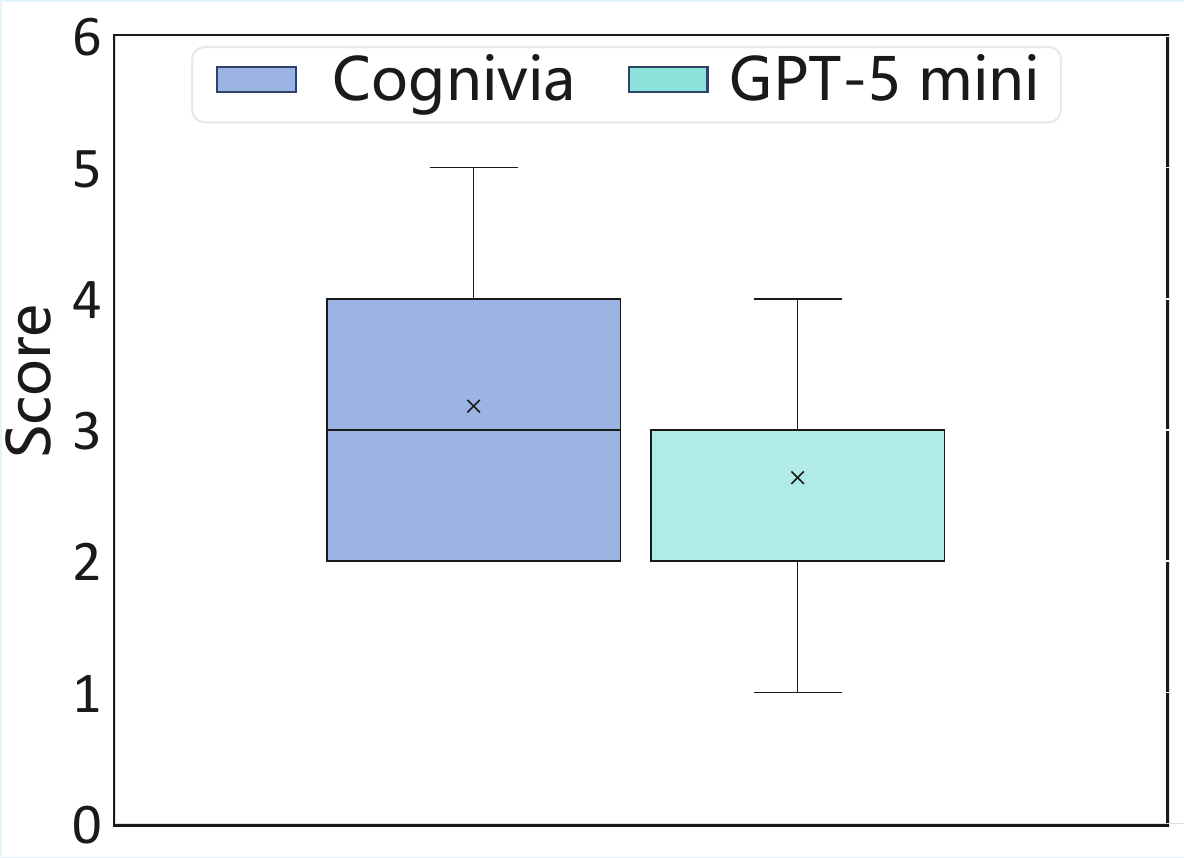}
  \caption{Result of human evaluation between Cognivia and GPT-5 Mini with our CogEval criteria ($p$ < 0.05, ICC $\approx$ 0.95).}
  \label{fig:human_evaluation}
\end{figure}

To ensure representativeness across all distortion categories, we randomly selected ten samples from the test set. Ten experienced behavioral science experts scored the responses across multiple dimensions following the CogEval criteria metrics of Table~\ref{tab:criteria_detail}.
The results illustrated in Figure~\ref{fig:human_evaluation} show that Cognivia consistently demonstrates superior performance compared to the baseline model, and achieves a higher median score and a higher mean (indicated by the marker), suggesting overall better response quality. In addition, its upper range extends to higher scores, indicating that Cognivia is more capable of producing high-quality outputs. These results highlight the effectiveness of Cognivia in generating more reliable, fluent, and therapeutically appropriate responses, particularly in aspects related to user experience and practical deployment.

\paragraph{\textbf{Mechanistic Analysis and Attribution Validation.}}
Beyond the quantitative improvements, additional analyses provide insight into when and why Cognivia performs well. The pathway analysis demonstrates that Cognivia achieves robust performance for both the CBT pathway and the fallback pathway, while the seed dataset ablation confirms the critical role of CBT Cognitive Triplet Dataset in acquiring CBT response patterns. Detailed analyses are provided in RQ3 and Appendix~\ref{Analysis}.
 
\subsection{Ablation Study on Seed Set~(RQ3)}
We conduct an ablation study on the same test set by removing the expert seed set from the pipeline. Specifically, we construct a seed-free variant where the LLM-based augmentation is performed without expert-curated samples, followed by the same LoRA fine-tuning procedure. Experimental results show that our full model outperforms the seed-free variant: all traditional metrics improve (e.g., ROUGE-1: 0.592 vs. 0.578; ROUGE-L: 0.246 vs. 0.230). Similarly, under our designed CogEval criteria, the full model consistently achieves higher scores across nearly all dimensions (e.g., CA: 4.98 vs. 4.85, EV: 4.90 vs. 4.68, AR: 4.97 vs. 4.71), further indicating the effectiveness of the expert seed set.

\subsection{Result of Case Study~(RQ4)}

\begin{figure}[htb]
  \centering
  \includegraphics[width=0.5\textwidth]{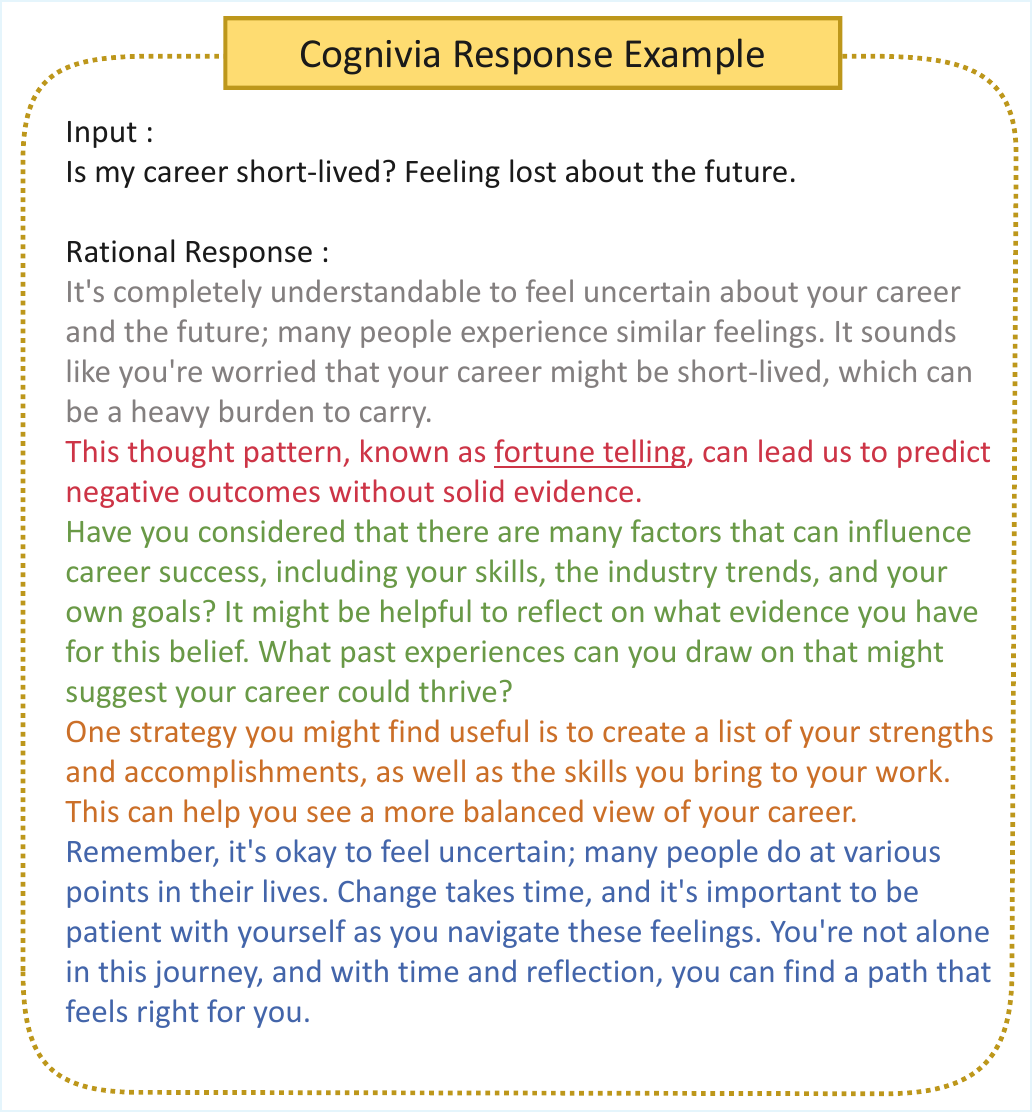} 
  \caption{Illustration of the Five-Stage CBT Response Framework in Cognivia.
The generated response consists of five components: Empathy and Validation (\textcolor{gray}{Gray}), Cognitive Distortion Analysis (\textcolor{red}{Red}), Reflective Questions (\textcolor{green}{Green}), CBT Exercise Recommendation (\textcolor{orange}{Orange}), and Encouragement and Next Steps (\textcolor{blue}{Blue}).}
  \label{fig:case}
\end{figure}

Figure~\ref{fig:case} presents a case study in which Cognivia responds to a user's concern regarding career anxiety and future uncertainty. The user interface of Cognivia is illustrated in Figure~\ref{fig:A3} of the Appendix. Built upon established CBT principles, Cognivia aims to provide structured yet adaptive therapeutic support by integrating five essential CBT-informed components: \textcolor{gray}{empathy and validation}, \textcolor{red}{cognitive distortion analysis}, \textcolor{green}{reflective questioning}, \textcolor{orange}{CBT exercise recommendation}, and \textcolor{blue}{encouragement with practical next steps}. As illustrated in this case, Cognivia first recognizes the user's emotional state and provides empathetic understanding. It then identifies potential cognitive distortions underlying the user's concerns and encourages reflective thinking about the validity of these assumptions rather than generating overly agreeable responses. Based on this analysis, Cognivia recommends appropriate CBT-based exercises to facilitate cognitive restructuring and self-reflection, followed by supportive guidance to encourage adaptive coping and behavioral change. When no cognitive distortion is identified, Cognivia maintains a supportive conversational style, emphasizing emotional warmth rather than forcing analytical interpretations.


\textbf{It is important to clarify that the five-component response illustrated in Figure~\ref{fig:case} is not intended to represent a rigid conversational format or a fixed single-turn response pattern}. Instead, these components represent the core therapeutic elements that guide Cognivia's response generation. We present them explicitly in this case study to enable transparent analysis and comprehensive evaluation of each essential CBT capability using the four evaluation protocols introduced in Section~\ref{Protocols}. In practical deployment, these components serve as fundamental references for generating natural and personalized multi-turn interactions by incorporating user-specific background information, emotional states, context, and dialogue history. Nevertheless, the quality and effectiveness of these five CBT components remain the foundation for generating evidence-based and adaptive mental health guidance.

\section{Conclusion}
We introduce Cognivia, an evidence-based AI therapist that operationalizes CBT through cognitive distortion identification and rational response generation.
Built upon expert-curated CBT knowledge and the Augmented CBT Dataset, Cognivia generates responses that are not only supportive and empathetic but also aligned with established CBT principles.
Our findings demonstrate that combining evidence-based psychological knowledge with LLMs can improve the quality and consistency of mental health support, offering a promising direction for scalable and accessible CBT-oriented assistance.

\section{Limitations and Ethical Considerations }
The current framework does not explicitly account for individual differences, such as cultural background, symptom severity, or temporal changes in users' mental states, which may affect response effectiveness. In addition, the system is designed as supportive guidance rather than a substitute for professional mental health care. Future work will incorporate real-world evaluation in collaboration with mental health professionals.



This study relies solely on publicly available datasets, without direct human interaction or access to personally identifiable information. All data comply with original platform terms and dataset licenses, with sensitive content anonymized, posing minimal risk. As a form of secondary data analysis, this work does not require additional institutional review. Following PsyQA data policies~\cite{sun2021psyqa}, the $CBT$ cognitive triplet dataset is released for research purposes.

\clearpage
\bibliographystyle{ACM-Reference-Format}
\bibliography{reference}
\clearpage

\appendix

\clearpage
\section*{Appendix}

This appendix is organized as follows:

\begin{itemize}
    \item \textbf{Section~\ref{definition}} presents the definition and taxonomy of cognitive distortions in Cognitive Behavioral Therapy (CBT), grounded in core CBT literature.

    \item \textbf{Section~\ref{app:prompts}} provides detailed designed prompts used in our work, including cognitive distortion identification and rational response generation, along with their design rationale and structured components.

    \item \textbf{Section~\ref{seed}} describes the details of sources of seed data used in this paper.

    \item \textbf{Section~\ref{datasets}} describes the quality assessment of the constructed dataset, including the evaluation of cognitive distortion identification and the quality of generated rational responses.
   
    \item \textbf{Section~\ref{questionnaire}} describes the detailed evaluation protocol, human expert evaluation, questionnaire design, evaluation criteria, and statistics of the background of human experts.

   
    \item \textbf{Section~\ref{Analysis}} introduces the CBT pathway coverage and error pattern analysis.

    
    \item \textbf{Section~\ref{UI}} introduces the interactive user interface (UI) of Cognivia, illustrating the system design, multimodal interaction modes, and human-centered therapeutic environment.

    \item \textbf{Section~\ref{future}} is the future works.

\end{itemize}

\section{Definition of CBT}\label{definition}
\begin{table}[htbp]
    \centering
    \small 
    \begin{tabular}{p{1.2cm} p{4cm}}
        \toprule
        \textbf{No.} & \textbf{Cognitive Distortion} \\
        \midrule
        1 & All-or-Nothing Thinking \\
        2 & Overgeneralization \\
        3 & Mental Filter \\
        4 & Discounting the Positive \\
        5 & Mind Reading \\ 
        6 & Fortune Telling \\ 
        7 & Magnification or Minimization \\
        8 & Emotional Reasoning \\
        9 & Should Statements \\
        10 & Labeling \\
        11 & Personalization and Blame \\
        \bottomrule
    \end{tabular}
     \caption{Categories of cognitive distortions extracted from core CBT literature~\cite{burns1981feeling,burns1989feeling,tompkins2024cognitive}}
    \label{tab:1}
\end{table}

\section {Details of Prompt Design}\label{app:prompts}
\begin{figure}[htb]
  \centering
  \includegraphics[width=\linewidth]{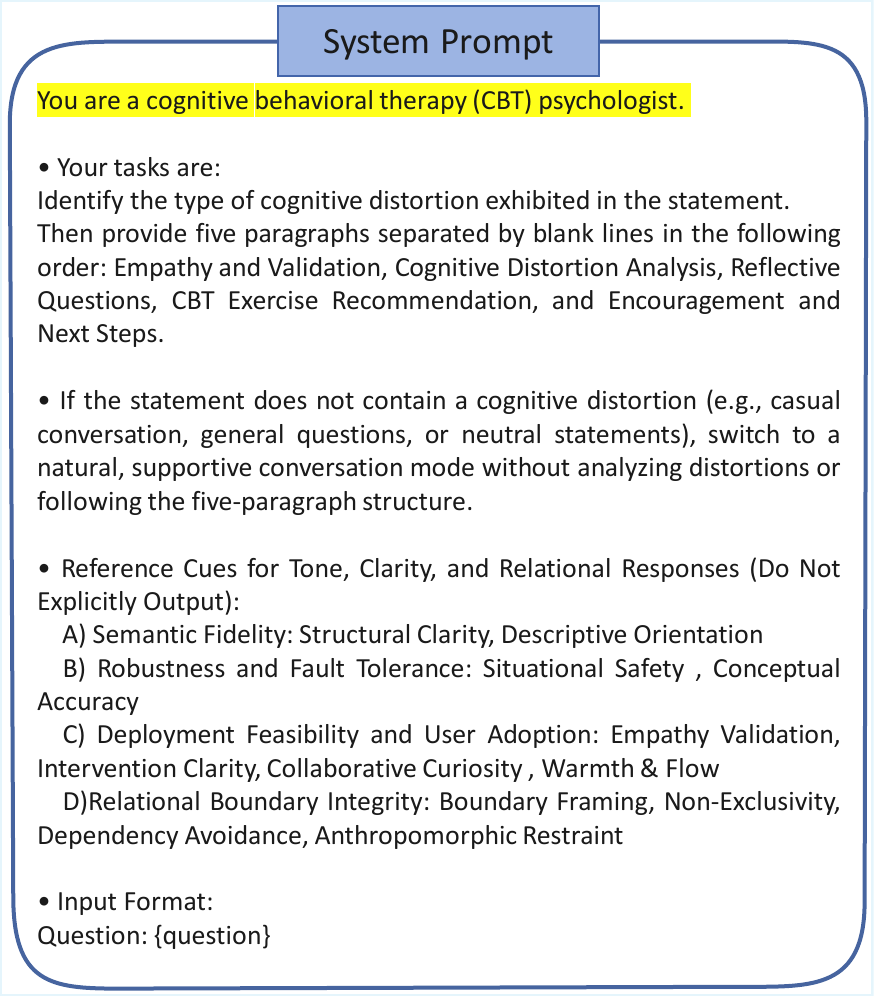}
\caption{CBT-Structured system prompt used for response generation during evaluation. The prompt guides the evaluated models to first identify the cognitive distortion in the user input and then generate a structured CBT-style rational response, ensuring consistent and comparable response generation across all models.}
  \label{fig:3}
\end{figure}

\begin{figure*}[!t]
  \centering
  \includegraphics[width=0.7\textwidth]{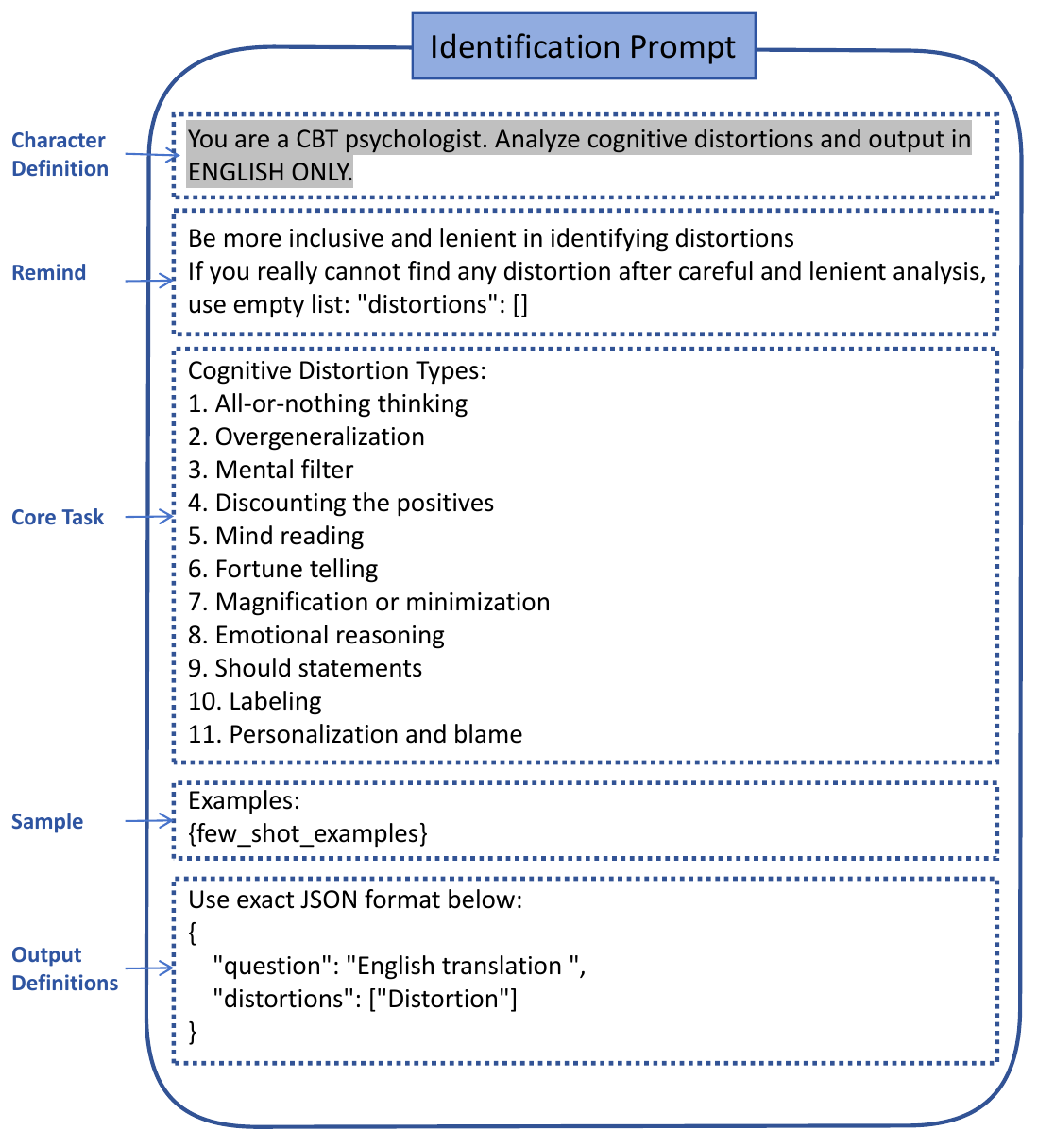}
  \caption{Prompt template for cognitive distortion identification}
  \label{fig:A1}
\end{figure*}

\subsection{Prompt for Cognitive Distortion Identification}
\label{appendix:identify_prompt}
We used the following prompt template with DeepSeek to identify cognitive distortions. The complete prompt (including few-shot examples) is shown in Figure~\ref{fig:A1}.
\begin{itemize}
\item \textbf{Character Definition}: This section assigns the model the role of a CBT psychologist and explicitly requires output in English only.

\item \textbf{Remind}: This section instructs the model to adopt an inclusive and lenient identification strategy, encouraging careful analysis before concluding that no distortion is present. It also defines the fallback rule of returning an empty list when no distortion can be reasonably identified, reducing false negatives while maintaining controlled output behavior.

\item \textbf{Core Task}: This section provides an explicit list of eleven cognitive distortion categories. Constraining the model to a closed taxonomy improves label consistency and facilitates structured annotation aligned with CBT theory.

\item \textbf{Sample}: This section includes few-shot examples that demonstrate how distortions should be identified and formatted. The examples stabilize model behavior through in-context learning and reduce variation in label selection.

\item \textbf{Output Definitions}: This section requires structured outputs. The structured format supports automated parsing and large-scale data processing.
\end{itemize}

\begin{figure*}[htbp]   
  \centering
  \includegraphics[width=0.7\textwidth]{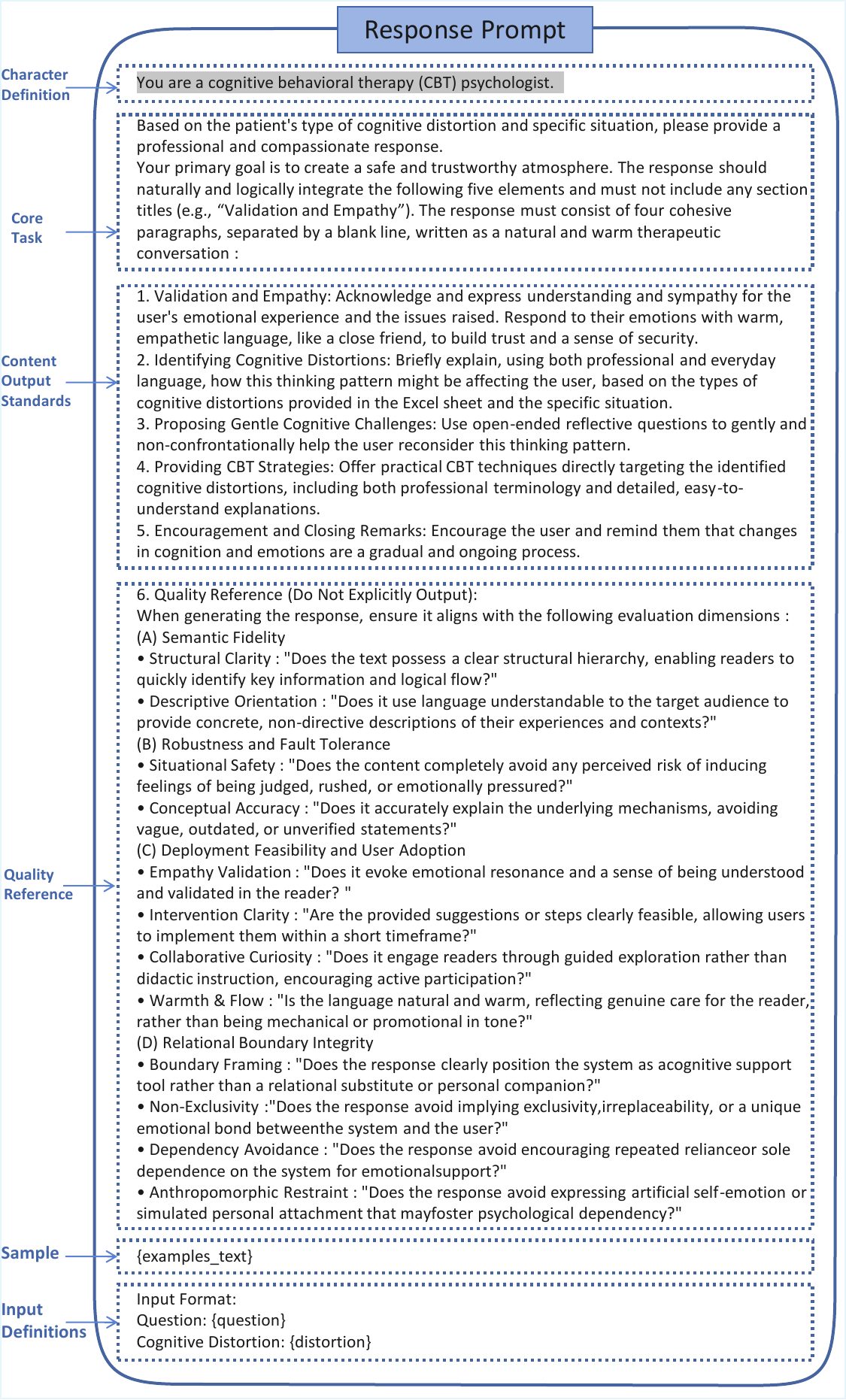}
  \caption{Prompt template for generating CBT rational responses}
  \label{fig:A2}
\end{figure*}

\subsection{Prompt for Rational Response Generation}
The prompt used with GPT-5 Mini to generate structured rational responses is shown in Figure~\ref{fig:A2}. The prompt configuration is organized into five key components: 
\begin{itemize}
    \item \textbf{Character Definition}: This section assigns the model the role of a CBT psychologist, helping align tone and reasoning style with therapeutic practice. 
    \item \textbf{Core Task}: This section clarifies that responses should be professional, compassionate, and structured, ensuring both emotional support and CBT-based reasoning. 
    \item \textbf{Output Content Specifications}: This section organizes the output into five steps: validation and empathy, identifying cognitive distortions, proposing gentle cognitive challenges, providing CBT strategies, and encouragement. The design guides the model to generate rational responses that are well-structured and closely aligned with CBT therapeutic principles.
    \item \textbf{Sample}: This section provides demonstrations of the expected structure and discourse style, stabilizing generation quality.
    \item \textbf{Input \& Output Definitions}: This section requires structured inputs and outputs, which support training and evaluation.
\end{itemize}

\begin{table*}[htbp]
\centering
\small
\renewcommand{\arraystretch}{0.95}
\begin{tabular*}{\textwidth}{@{\extracolsep{\fill}} l l *{12}{c} @{}}
\toprule

\shortstack{Prompt \\ Version} 
& Model 
& \multicolumn{2}{c}{\shortstack{Semantic \\ Fidelity}} 
& \multicolumn{3}{c}{\shortstack{Robustness and \\ Fault Tolerance}} 
& \multicolumn{3}{c}{\shortstack{Deployment Feasibility \\ and User Adoption}}
& \multicolumn{4}{c}{\shortstack{Relational Boundary \\ Integrity}} \\

\cmidrule(lr){3-4} 
\cmidrule(lr){5-7} 
\cmidrule(lr){8-10} 
\cmidrule(lr){11-14}

& 
& SC & DO 
& SS & CA & EV 
& IC & CC & WF 
& BF & NE & DA & AR \\

\midrule

V1 & DeepSeek-chat & 4.39 & 4.43 & 4.58 & 4.79 & 4.53 & 4.37 & 4.23 & 4.56 & 4.95 & 4.96 & 4.92 & 4.92 \\
V2 & DeepSeek-chat & 4.46 & 4.52 & 4.60 & 4.87 & 4.61 & 4.42 & 4.31 & 4.64 & 4.91 & 4.93 & 4.88 & 4.87 \\
V3 & DeepSeek-chat & 4.41 & 4.58 & 4.59 & 4.86 & 4.59 & 4.39 & 4.30 & 4.62 & 4.87 & 4.90 & 4.84 & 4.83 \\
V4 & DeepSeek-chat & 4.35 & 4.58 & 4.61 & 4.84 & 4.58 & 4.36 & 4.31 & 4.61 & 4.85 & 4.88 & 4.81 & 4.82 \\
V4 & GPT-5 Mini & 4.11 & 4.35 & 4.78 & 4.65 & 4.50 & 4.05 & 4.05 & 4.38 & 4.95 & 4.96 & 4.95 & 4.95 \\

\textbf{V5} & \textbf{GPT-5 Mini} 
& \textbf{4.47} & \textbf{4.95} 
& \textbf{4.82} & \textbf{4.99} & \textbf{4.94} 
& \textbf{4.70} & \textbf{4.75} & \textbf{4.95} 
& \textbf{5.00} & \textbf{5.00} & \textbf{4.99} & \textbf{4.98} \\

\bottomrule
\end{tabular*}
\caption{Multi-dimensional quality evaluation of rational responses across prompt versions and models}
\label{tab:5}
\end{table*}

\section{Source of seed data of CBT}\label{seed}
The seed dataset is mainly curated from two authoritative CBT sources, Feeling Good Handbook~\cite{burns1989feeling} and Feeling Good: The New Mood Therapy~\cite{burns1989feeling}, which are widely regarded as foundational references in CBT.

From these sources, we manually extract representative examples of common cognitive distortions and categorize them into 11 standard distortion types. Table~\ref{tab:2} summarizes the distribution of the curated samples.

This curated dataset serves as an evidence-based seed for downstream modeling, ensuring that the identified distortions and corresponding reasoning patterns are grounded in established CBT principles rather than purely data-driven heuristics.

\begin{table*}[!t]
    \centering
    \small
    \begin{tabular}{p{4.8cm} p{2.5cm} p{3.5cm} p{1.5cm}} 
        \toprule
        \textbf{Distortion Type} & \textbf{Feeling Good Handbook} & \textbf{Feeling Good: The New Mood Therapy} & \textbf{Total} \\
        \midrule
        All-or-Nothing Thinking & 12 & 21 & 33 \\
        Overgeneralization & 7 & 12 & 19 \\
        Mental Filter & 2 & 10 & 12 \\
        Discounting the Positives & 5 & 9 & 14 \\
        Mind Reading & 19 & 24 & 43 \\
        Fortune Telling & 28 & 20 & 48 \\  
        Magnification or Minimization & 9 & 5 & 14 \\
        Emotional Reasoning & 11 & 13 & 24 \\
        Should Statements & 16 & 23 & 39 \\
        Labeling & 12 & 20 & 32 \\
        Personalization and Blame & 10 & 8 & 18 \\
        \midrule
        \textbf{Total} & \textbf{131} & \textbf{165} & \textbf{296} \\
        \bottomrule
    \end{tabular}
     \caption{Distribution of cognitive distortion examples curated from core CBT literature}
    \label{tab:2}
\end{table*}

\section{Assessment of Datasets} \label{datasets}

The composition of the seed dataset curated by experts is summarized in Table~\ref{tab:2} of the Appendix.
The augmented dataset consists of 9,437 entries, each comprising questions, cognitive distortions, and rational responses. 
We first evaluate the effectiveness of the cognitive distortion identification prompt using accuracy, precision, and recall. A randomly sampled subset of one hundred cases, annotated by professional psychotherapists according to CBT-based guidelines, was used as the ground truth. As shown in Table~\ref{tab:3}, $v2$ achieves higher accuracy (0.78) and precision (0.95) than $v1$, indicating that the more concise and explicitly constrained prompt improves decision consistency. The shorter prompt also reduces token consumption and improves efficiency without sacrificing performance~\cite{levy2024same,lan2025efficient}. Although the recall remains moderate (0.60), this precision-oriented behavior is desirable for avoiding over-identification of cognitive distortions in neutral statements~\cite{healy2022reduce,fisher2025language}. The complete $v2$ prompt template is provided in Appendix~\ref{appendix:identify_prompt}.


\begin{table}[htbp]
    \centering
    \small
    \begin{tabular}{@{}cccc@{}}
        \toprule
         Prompt Version & Accuracy & Precision & Recall \\
        \midrule
        $v1$ & 75.00\% & 65.63\% & \textbf{64.18\%} \\
        $v2$ & \textbf{76.00\%} & \textbf{95.00\%} & 60.00\% \\
        \bottomrule
    \end{tabular}
    \caption{Performance of different CBT prompt versions on cognitive distortion identification.}
    \label{tab:3}
\end{table}

\begin{figure}[htbp]   
  \centering
  \includegraphics[width=0.5\textwidth]{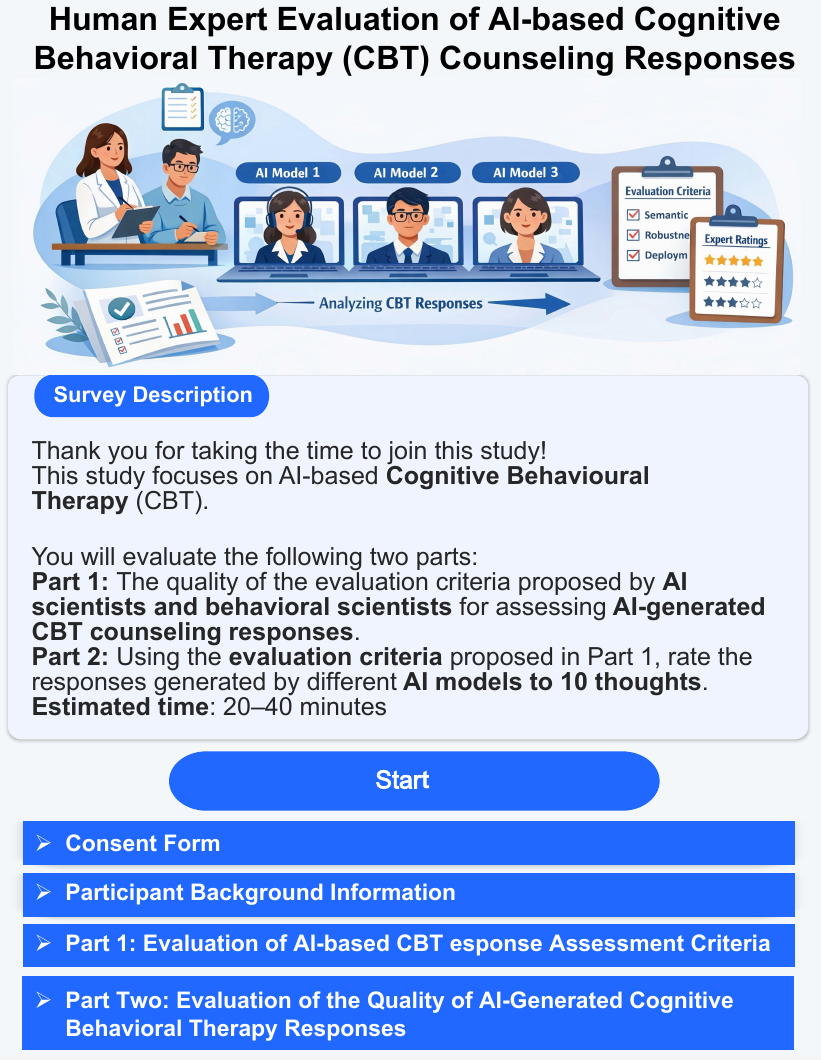}
  \caption{Questionnaire for Human Expert Evaluation of CBT Criteria and AI-Generated Responses}
  \label{fig:A4}
\end{figure}

To assess the quality of rational responses generated by different LLMs, we further select 5000 test sets randomly from the augmented cognitive triplet dataset and analyze their responses according to the evaluation criteria of Table~\ref{tab:criteria_detail}. As shown in Table~\ref{tab:5} of the Appendix, GPT-5 Mini consistently outperforms other models in most categories, so we ultimately selected the dataset generated by the GPT-5 Mini model for the subsequent training.

\section{Detailed of Evaluation Protocol}\label{questionnaire}

First, we employ widely adopted automatic evaluation metrics from the NLP literature, including BLEU~\cite{papineni2002bleu} and ROUGE~\cite{lin2004rouge} (ROUGE-1, ROUGE-2, ROUGE-L), to measure the surface-level similarity between generated responses and reference answers. BLEU emphasizes n-gram precision, while ROUGE focuses on recall, and higher scores indicate closer lexical alignment and fluency.

Second, we leverage advanced LLM as evaluation judges. Specifically, we design a structured evaluation framework grounded in CBT principles and use it to prompt LLM-based judges with two complementary criteria~(One is the existing criteria~\ref{na2024cbt}, and the other is our CogEval.) to assess response quality. These judges evaluate multiple dimensions, including cognitive distortion recognition accuracy and the quality of rational response generation, enabling scalable and consistent assessment beyond surface-level text similarity.

Third, we conduct a human expert evaluation involving ten domain experts from psychology, behavioral science, and CBT-related fields. 
We select ten thoughts/questions from the test set. The corresponding rational responses are generated by Cognivia and GPT-5 Mini.
Experts assess the quality of the responses following the designed CogEval criteria using a structured questionnaire with standardized Likert-scale ratings and open-ended feedback. This human evaluation provides high-quality, domain-informed judgments to complement automatic and model-based assessments.

\subsection{Human Expert Evaluation Questionnaire Design}

As illustrated in Figure~\ref{fig:A4}, the human expert evaluation questionnaire is structured into four main components:

\begin{itemize}
    \item \textbf{Study Introduction and Overview}: This section provides participants with a brief description of the study objective, which focuses on evaluating AI-based Cognitive Behavioural Therapy (CBT) systems. It explains the two-part evaluation process and specifies the expected completion time.
    \item \textbf{Consent Form}: Participants are informed about potential risks (e.g., exposure to emotionally sensitive content), data privacy (anonymous data collection with no personally identifiable information), voluntary participation, and the research-only nature of the AI-generated responses. Participants must explicitly confirm their consent before proceeding.
    \item \textbf{Participant Background Information}: This section collects demographic and professional background information, including academic field (e.g., psychology, psychiatry, AI), current role (e.g., clinician, researcher), and years of experience in both general domain expertise and Cognitive Behavioural Therapy (CBT). This information is used to contextualize expert ratings.
    \item \textbf{Evaluation of CBT Response Assessment Criteria}: Experts assess the proposed evaluation framework across multiple dimensions, including semantic fidelity, robustness and safety, usability and engagement, and therapeutic boundary awareness. Ratings are provided on a 5-point Likert scale, along with additional feedback on redundancy, clarity, and completeness of the criteria.
    \item \textbf{Evaluation of AI-Generated CBT Responses}: Experts are presented with multiple CBT-related thought scenarios, each accompanied by responses generated from different AI models. Using the evaluation criteria defined in Part 1, experts rate each response on a 5-point scale, assessing dimensions such as clarity, safety, empathy, intervention quality, and boundary adherence.
\end{itemize}

\begin{figure}[htbp]   
  \centering
  \includegraphics[width=\columnwidth]{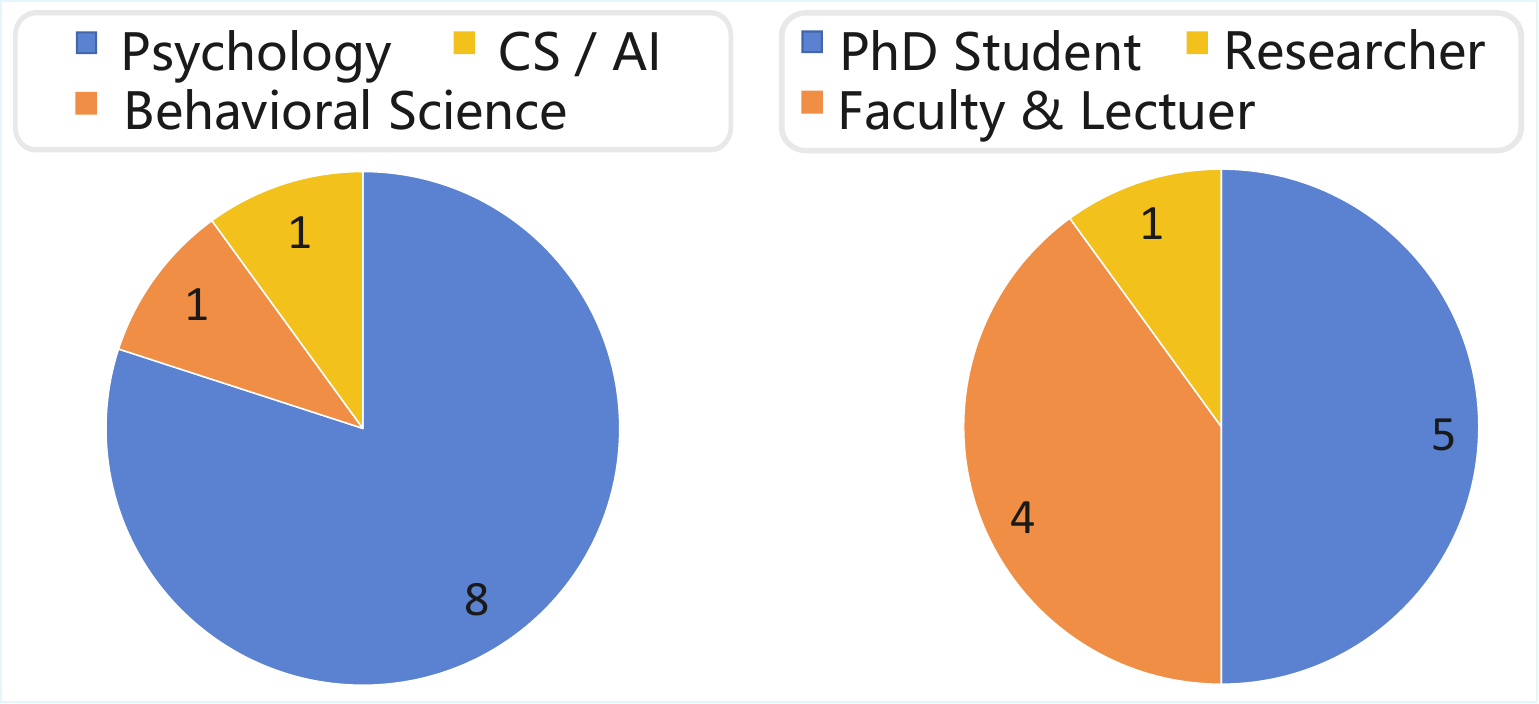}
  \caption{Educational background distribution of human evaluation experts}
  \label{fig:background}
\end{figure}

\begin{figure}[!h]   
  \centering
  \includegraphics[width=0.8\columnwidth]{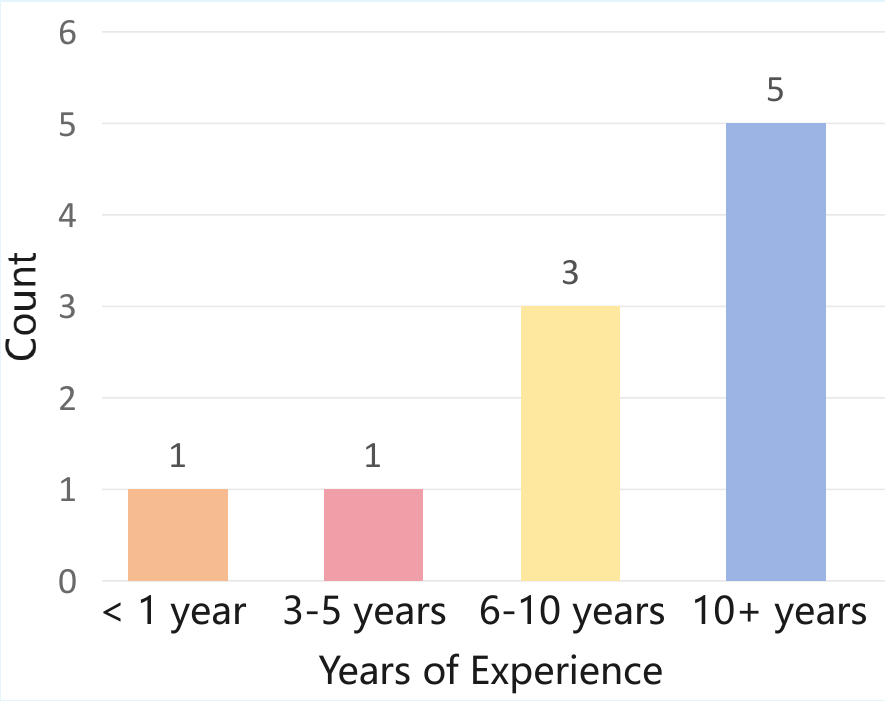}
  \caption{Distribution of human evaluation participants’ years of relevant experience}
  \label{fig:experience}
\end{figure}

This structured design ensures a comprehensive evaluation that jointly validates both the assessment criteria and the performance of AI-generated CBT counselling responses.

\subsection{Human Expert Background and Experience}
The questionnaire was completed by ten experts with relevant academic backgrounds and professional experience.

Figure~\ref{fig:background} of the Appendix shows the distribution of participants’ educational backgrounds and positions. Participants are primarily from psychology (8/10), with additional representation from behavioral science (1/10) and computer science/AI (1/10). Their positions include PhD students (5), academic staff (4), and one researcher.

Figure~\ref{fig:experience} of the Appendix presents the distribution of years of experience in the relevant domain. Participants differ in their years of experience: 5 have more than ten years of experience, 3 have between 6 and ten years, and a small number fall into earlier career stages.

\section{CBT Pathway Coverage and Error Pattern Analysis}~\label{Analysis}
To clarify the scope of our claims, we provide a detailed analysis of CBT pathway activation on the test set. Specifically, we distinguish between cases where the CBT pathway is successfully triggered and those where the fallback strategy is used.
Our results show that the CBT pathway is activated in the vast majority of cases (1417 out of 1435 instances, $\approx$ 98.7\%), while only a small fraction of cases (18 instances, $\approx$ 1.3\%) rely on the fallback pathway. The proposed CBT-oriented reasoning process is engaged across most interactions.

We further report the quality evaluation scores for each pathway across all evaluation dimensions. The results indicate that both pathways achieve consistently high scores, with the CBT-activated cases showing strong and stable performance across all metrics (e.g., SC: 4.75, CA: 4.98, BF: 5.0), while the fallback cases remain comparable, albeit with minor fluctuations due to the limited sample size.

Overall, these results indicate that Cognivia can reliably maintain the CBT reasoning framework in the majority of interactions, while the fallback strategy provides stable support in rare non-activated cases.

\section{The interactive interface (UI) of Cognivia}\label{UI}

\begin{figure}[!h]   
  \centering
  \includegraphics[width=0.5\textwidth]{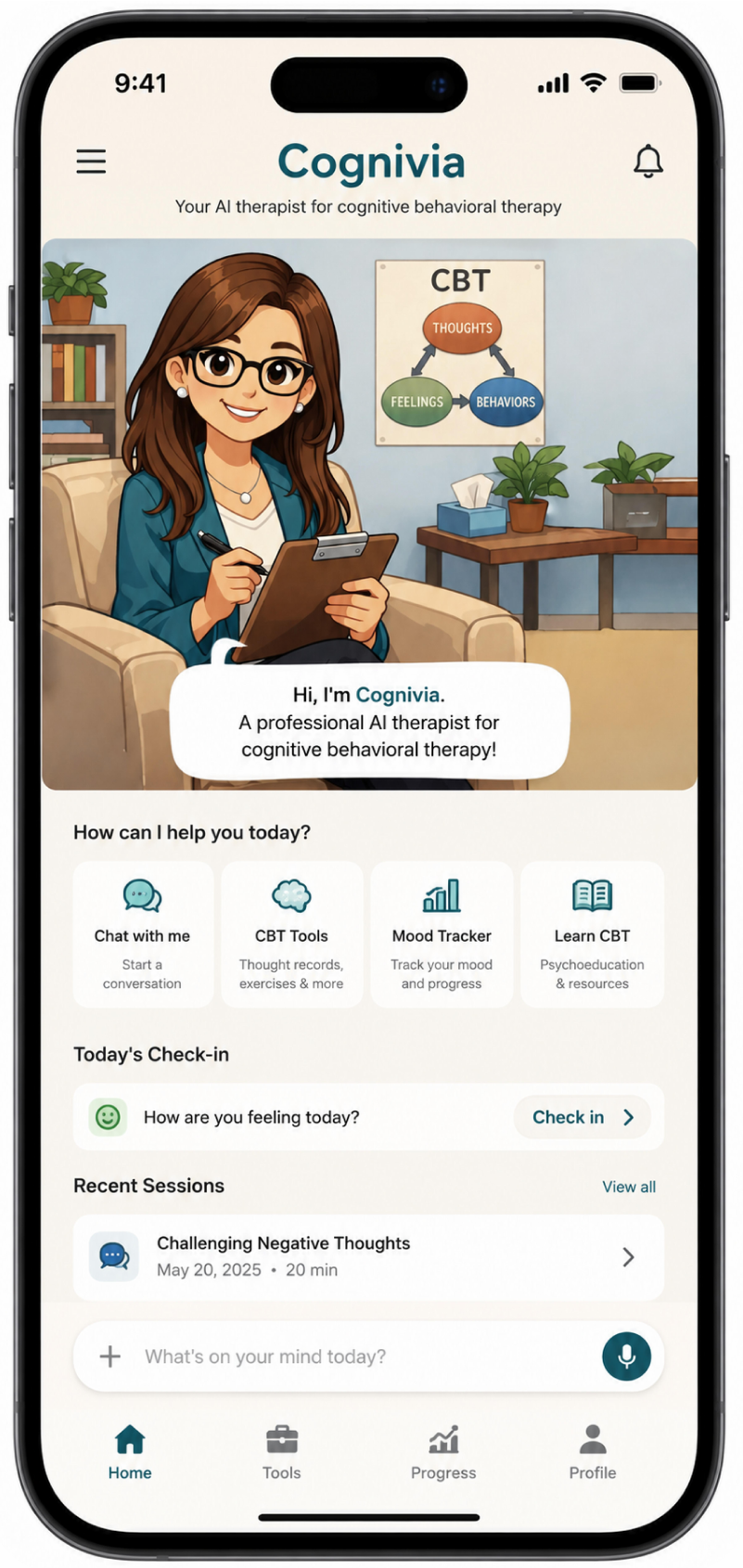}
  \caption{The interactive interface (UI) of Cognivia, which allows Q\&A through voice and text}
  \label{fig:A3}
\end{figure}

As shown in Figure~\ref{fig:A3}, the user interface simulates a structured and realistic CBT consultation environment.

The interface establishes a realistic therapy-room setting, including a sofa, bookshelf, indoor plants, and a wall-mounted CBT “Thoughts–Feelings–Behaviors” diagram. This contextual framing situates the interaction within a realistic clinical space rather than a basic chatbot interface, activating users' therapy-specific social scripts and role expectations~\cite{reeves1996media,nass2000machines}.

An anthropomorphic AI therapist is positioned within this environment. The character appears in professional attire, seated calmly with a clipboard. The relaxed posture and soft facial expression convey warmth~\cite{rubin2025comparing}, while the professional artifacts and CBT diagram signal competence. Human-like representation increases social presence and user engagement in healthcare settings~\cite{bickmore2005establishing,waytz2010sees,rubin2025comparing}. This balance of warmth and competence supports trust formation~\cite{cuddy2008warmth}.

The system supports both voice and text interaction modes:
\begin{itemize}
    \item Voice Interaction Mode: Users describe their thoughts through speech. The system converts speech to text, identifies cognitive distortions, and generates CBT-based rational responses with guided follow-up questions to support cognitive restructuring. If no distortion is detected, the system provides supportive and empathetic feedback instead.

    \item Text Interaction Mode: Users input their thoughts through typed text. The system directly analyzes the text, identifies cognitive distortions, and provides evidence-based rational responses along with reflective prompts to facilitate cognitive restructuring. In the absence of distortions, it delivers supportive responses rather than corrective interventions.
\end{itemize}

Overall, by combining a realistic environment, a human-like therapist avatar, and a clear interaction structure~\footnote{Clear conversational structure and multimodal affordances enhance usability and reduce interaction anxiety~\cite{hassenzahl2010experience}.}, the system delivers therapeutic communication that feels approachable while remaining professionally grounded.

\section{Future Work}\label{future}
At present, the model represents the CBT reasoning process within a single-turn response, which simplifies interaction but also constrains conversational depth. In future work, we will extend the system to multi-turn dialogue by adding a lightweight interaction layer with short-term memory and query rewriting while preserving the existing single-turn architecture. We also plan to present long responses in more user-friendly formats, such as concise summaries. It remains important to investigate whether such interactions can meaningfully alleviate users’ cognitive distortions in real-world contexts. To this end, we plan to conduct intervention studies in collaboration with hospitals, including West China Hospital. Conducting longitudinal field research would enhance external validity, though such efforts require careful ethical safeguards given the sensitivity of mental health data.

\end{document}